\documentclass[10pt]{article}
\usepackage[margin=1.3in]{geometry}
\usepackage{amsmath,amssymb,amsthm}
\usepackage[authoryear,round,sort]{natbib}
\usepackage[colorlinks=true,linkcolor=blue,citecolor=blue,urlcolor=blue]{hyperref}

\usepackage{setspace}
\usepackage{etoolbox}
\makeatletter
\preto\proof{\setlength{\topsep}{2pt}\setlength{\parsep}{1pt}}
\makeatother

\usepackage[compact]{titlesec}

\theoremstyle{plain}
\newtheorem{theorem}{Theorem}[section]
\newtheorem{lemma}[theorem]{Lemma}
\newtheorem{proposition}[theorem]{Proposition}
\theoremstyle{definition}
\newtheorem{definition}[theorem]{Definition}
\newtheorem{remark}[theorem]{Remark}

\newcommand{\R}{\mathbb{R}}
\newcommand{\Qstruct}{\mathcal{Q}_{\mathrm{struct}}}
\newcommand{\sgn}{\operatorname{sgn}}
\newcommand{\Rank}{\operatorname{Rank}}
\newcommand{\norm}[1]{\left\|#1\right\|}

\title{Structural Incompatibility of Differentiable Sorting\\
and Within-Vector Rank Normalization}
\author{Taeyun Kim\\
\small College of Liberal Studies, Seoul National University, Seoul, Republic of Korea\\
\small \texttt{auntymike@snu.ac.kr}}
\date{}

\begin{document}
\maketitle

\begin{abstract}
We show that differentiable sorting and ranking operators are structurally
incompatible with within-vector rank normalization. We formalize admissibility
through monotone invariance (C1), batch independence (C2), and a rank-space
stability condition (C3). Gap-sensitive relaxations such as SoftSort violate (C1)
by a quantitative margin that depends on the temperature and input scale. Batchwise
rank relaxations such as SinkhornSort violate (C2): the same sample can be assigned
outputs arbitrarily close to $0$ or $1$ depending solely on batch context.
Condition (C3) implies (C1) under the rank representation used here and should not
be read as a third independent failure mode. We also characterize the admissible
class: any admissible operator must factor through the rank representation via a
Lipschitz function.
\end{abstract}

\section{Introduction}
\label{sec:intro}
Many learning tasks operate on vectors whose coordinates represent scores, rankings,
or preference magnitudes, as in information retrieval, recommendation, and
multi-criteria evaluation. When the absolute scale of each coordinate is
uninformative, a normalization operator should be invariant under strictly monotone
rescalings, independent of mini-batch composition, and stable under small
perturbations.

Differentiable sorting and ranking
methods~\citep{Blondel2020,Cuturi2019,Grover2019,Petersen2021,Prillo2020}
construct smooth surrogates for sorting or permutation-valued outputs in gradient-based
optimization. They are sometimes used as rank-based input representations. We show
that this use is structurally inadmissible: the design choices that make these operators
useful as sorting surrogates can conflict with the requirements of within-vector rank
normalization.

We introduce three admissibility conditions
(C1)--(C3) and the class $\Qstruct$ of within-vector rank normalizers
(\S\ref{sec:setup}). We establish quantitative lower bounds on the (C1)
violation for SoftSort-type operators (Proposition~\ref{prop:c1}) and on the (C2)
violation for batchwise rank relaxations (Proposition~\ref{prop:c2}). We show that
(C3) implies (C1) under the present rank representation and therefore does not
constitute an independent failure mode (Proposition~\ref{prop:c3_implies_c1}). We
characterize $\Qstruct$ via a factorization theorem (Theorem~\ref{thm:structural_form}).

Classical nonparametric statistics and copula theory use rank invariance for
estimation~\citep{Hajek1967,Huber1964,Nelsen2006}. Learning-to-rank optimizes
rankings as relational outputs over batches~\citep{Burges2005,Cao2007,Joachims2002}.
Lipschitz stability of networks has been studied in value
space~\citep{Fazlyab2019,Gouk2021,Scaman2018}. To our knowledge, this literature
does not treat within-vector rank normalization as an axiomatic operator-design problem.

Recent advances refine differentiable ordering surrogates through monotonic sorting
networks, generalized neural sorting networks with differentiable swap functions, and
unified constructions for ranking, sorting, and top-$k$
selection~\citep{Kim2024,Petersen2022,Struski2025}. Our question is orthogonal to these
optimization-oriented improvements. The issue here is not surrogate quality, but
admissibility as a within-vector rank normalizer.
\section{Setup}
\label{sec:setup}
\paragraph{Order equivalence.}
Let $g : \R \to \R$ be strictly increasing and write $g^{\otimes d}(x) :=
(g(x_1),\ldots,g(x_d))$. Two tie-free inputs $x, x' \in \R^d$ are
\emph{order-equivalent}, $x \sim_{\mathrm{ord}} x'$, if
\begin{equation}
  \sgn(x_i - x_j) = \sgn(x'_i - x'_j) \qquad \forall\, i \neq j.
  \label{eq:order_equiv}
\end{equation}
We restrict to a common scalar $g$ rather than coordinate-wise functions $g_i$
because coordinate-wise transformations do not generally preserve inter-coordinate
comparisons and so fall outside the invariance structure considered here.

\paragraph{Rank representation.}
For tie-free $x \in \R^d$, define the \emph{within-vector rank representation}
$\rho : \R^d \to \{0, \tfrac{1}{d-1}, \ldots, 1\}^d$ by
\begin{equation}
  \rho_i(x) := \frac{1}{d-1}\sum_{j \neq i} \mathbf{1}[x_j < x_i].
  \label{eq:rho}
\end{equation}
By construction, $\rho(x) = \rho(x')$ whenever $x \sim_{\mathrm{ord}} x'$, and
$\rho(g^{\otimes d}(x)) = \rho(x)$ for every strictly increasing $g$. The image
$\mathcal{R} \subset \{0,\frac{1}{d-1},\ldots,1\}^d$ is finite, comprising the rank
patterns achievable by tie-free inputs.

\paragraph{Admissibility.}
When an operator is evaluated relative to an ambient batch context $B$, we write
its output as $Q(x \mid B)$, where $x$ is the distinguished sample and $B$ is
held fixed.

\begin{definition}[Admissible operator]
\label{def:admissible}
An operator $Q(x \mid B) \in [0,1]$, defined on tie-free inputs, is \emph{admissible}
if it satisfies the following three conditions.

\medskip\noindent
\textbf{(C1) Monotone invariance.} For every strictly increasing scalar $g$,
\begin{equation}
  Q(g^{\otimes d}(x) \mid B) = Q(x \mid B) \qquad \forall\, x, B.
  \label{eq:C1}
\end{equation}

\medskip\noindent
\textbf{(C2) Batch independence.} For any two batches $B_1, B_2$ both containing $x$,
\begin{equation}
  Q(x \mid B_1) = Q(x \mid B_2).
  \label{eq:C2}
\end{equation}
We write $Q(x)$ when this holds.

\medskip\noindent
\textbf{(C3) Lipschitz stability.} There exists $L < \infty$ such that
\begin{equation}
  |Q(x \mid B) - Q(x' \mid B)| \le L\,\norm{\rho(x) - \rho(x')}
  \qquad \forall\, x, x', B.
  \label{eq:C3}
\end{equation}
\end{definition}

The admissible class is $\Qstruct := \{Q : Q \text{ satisfies (C1), (C2), (C3)}\}$.
\section{Structural Incompatibility}
\label{sec:incompat}
We show that SoftSort-type and batchwise rank operators are not in $\Qstruct$.
For (C1) and (C2) we give quantitative lower bounds on the violation.
Proposition~\ref{prop:c3_implies_c1} clarifies that (C3) implies (C1) here,
so there are two independent failure modes, not three.

Pointwise differentiable argsort relaxations construct weights through kernels
$\exp(-\tau(x_i - x_j)^2)$ that depend explicitly on value gaps $x_i - x_j$.
SoftSort \citep{Prillo2020} is the canonical example. More recent differentiable
sorting-network variants pursue alternative smooth parameterizations of local swap
operations~\citep{Kim2024,Petersen2022}, but the present proposition isolates,
in a minimal gap-sensitive example, the dependence on value magnitudes that conflicts
with (C1).

\begin{proposition}
\label{prop:c1}
Let $d = 2$, $\tau > 0$, and consider
\begin{equation}
  Q_1^\tau(x_1, x_2)
  := \frac{\exp(-\tau(x_1-x_2)^2)}{1 + \exp(-\tau(x_1-x_2)^2)}.
\end{equation}
Fix $\delta > 0$, let $x = (\delta, 0)$, and $g_M(t) = Mt$ with $M > 1$.
Then $g_M^{\otimes 2}(x) \sim_{\mathrm{ord}} x$ yet
\begin{equation}
  \bigl|Q_1^\tau(g_M^{\otimes 2}(x)) - Q_1^\tau(x)\bigr|
  = \bigl|\sigma(-\tau M^2\delta^2) - \sigma(-\tau\delta^2)\bigr| > 0,
  \label{eq:c1_violation}
\end{equation}
where $\sigma(t) = 1/(1+e^{-t})$. Hence $Q^\tau$ fails \textnormal{(C1)}.
\end{proposition}

\begin{proof}
Since $g_M$ is strictly increasing, $g_M^{\otimes 2}$ preserves $x_1 > x_2$, so
$\rho(g_M^{\otimes 2}(x)) = \rho(x)$. The scaled gap is $M\delta$:
\[
  Q_1^\tau(x) = \sigma(-\tau\delta^2), \qquad
  Q_1^\tau(g_M^{\otimes 2}(x)) = \sigma(-\tau M^2\delta^2).
\]
Since $M > 1$ and $\delta > 0$, we have $M^2\delta^2 > \delta^2$, and strict
monotonicity of $\sigma$ gives the violation in~\eqref{eq:c1_violation}.
\end{proof}

\begin{remark}
For $\tau = 1$, $\delta = 1$, $M = 2$: $|\sigma(-4) - \sigma(-1)| \approx 0.251$.
The argument extends to any kernel $\exp(-\tau\phi(\Delta))$ with $\phi$ strictly
convex and $\phi(0) = 0$. It is also not specific to $d = 2$: for general $d$,
fix any $i \neq j$ with $x_i > x_j$ and apply the same scaling; the gap
$x_i - x_j$ becomes $M(x_i - x_j)$, and the same monotonicity argument gives
a positive violation.
\end{remark}

Batchwise differentiable rank operators return the rank of each element relative to
the surrounding batch. This includes NeuralSort \citep{Grover2019}, SinkhornSort
\citep{Cuturi2019}, and the permutahedron-projection approach \citep{Blondel2020}.

\begin{proposition}
\label{prop:c2}
Let $Q^\lambda(z \mid B) \in [0,1]$ be a differentiable batchwise rank operator
normalized so that the lowest rank maps to $0$ and the highest to $1$. Assume that
for every tie-free batch $B$,
\begin{equation}
  Q^\lambda(z \mid B) \to \Rank_B(z) \quad \text{as } \lambda \to \infty,
\end{equation}
where $\Rank_B(z) \in [0,1]$ is the normalized hard rank of $z$ in $B$.
For any $\Delta > 1$ and $0 < \eta < 1$, let
\begin{equation}
  B^- := \{-\Delta,\,-\Delta+\eta,\,0\}, \qquad
  B^+ := \{0,\,\Delta,\,\Delta+\eta\}.
\end{equation}
Both contain $z = 0$. Then $\Rank_{B^-}(0) = 1$ and $\Rank_{B^+}(0) = 0$, and for
every $\varepsilon \in (0, 1/2)$ there exists $\Lambda$ such that for all
$\lambda \ge \Lambda$,
\begin{equation}
  \bigl|Q^\lambda(0 \mid B^-) - Q^\lambda(0 \mid B^+)\bigr| \ge 1 - 2\varepsilon.
  \label{eq:c2_violation}
\end{equation}
In particular, \textnormal{(C2)} fails.
\end{proposition}

\begin{proof}
In $B^-$ the element $0$ is the strict maximum (since $\Delta > 1 > \eta$), so
$\Rank_{B^-}(0) = 1$; in $B^+$ it is the strict minimum, so $\Rank_{B^+}(0) = 0$.
By the consistency assumption, $Q^\lambda(0 \mid B^-) \to 1$ and
$Q^\lambda(0 \mid B^+) \to 0$. Choosing $\Lambda$ so both outputs are within
$\varepsilon$ of their limits gives~\eqref{eq:c2_violation}.
\end{proof}

\begin{remark}
The consistency assumption is satisfied by the canonical examples cited above.
Ties are avoided by $\eta > 0$.
\end{remark}

We turn to (C3). The image $\mathcal{R}$ of $\rho$ is finite,
since tie-free inputs in $\R^d$ admit only finitely many rank patterns.

\begin{proposition}
\label{prop:c3_implies_c1}
If $Q$ satisfies \textnormal{(C3)}, then $Q$ also satisfies \textnormal{(C1)}.
\end{proposition}

\begin{proof}
Let $x \sim_{\mathrm{ord}} x'$. Then $\rho(x) = \rho(x')$, so
$\norm{\rho(x) - \rho(x')} = 0$. Applying (C3):
$|Q(x \mid B) - Q(x' \mid B)| = 0$ for all $B$. Since
$g^{\otimes d}(x) \sim_{\mathrm{ord}} x$ for every strictly increasing $g$,
(C1) follows.
\end{proof}

\begin{remark}
\label{rem:c3_role}
Because $\mathcal{R}$ is finite, any $H : \mathcal{R} \to [0,1]$ is automatically
Lipschitz: distinct points in $\mathcal{R}$ are at least some $\delta_* > 0$ apart,
so $L = 1/\delta_*$ suffices. Once an operator factors through $\rho$ by (C1),
condition (C3) adds at most a mild regularity requirement. Its role here is to
connect admissibility to a metric structure on $\mathcal{R}$ and to provide a clean
condition in Theorem~\ref{thm:structural_form}, not to serve as a structural axis
independent of (C1).
\end{remark}

\begin{theorem}[Structural incompatibility]
\label{thm:incompat}
SoftSort-type gap-sensitive relaxations and batchwise differentiable rank operators
do not belong to $\Qstruct$.
\end{theorem}

\begin{proof}
By Proposition~\ref{prop:c1}, SoftSort-type operators fail (C1). By
Proposition~\ref{prop:c2}, batchwise rank operators fail (C2). By
Proposition~\ref{prop:c3_implies_c1}, (C3) implies (C1), so any operator failing
(C1) also fails (C3). Neither family belongs to $\Qstruct$.
\end{proof}
\section{The Admissible Class}
\label{sec:structure}
\begin{theorem}[Representation form]
\label{thm:structural_form}
Let $Q$ satisfy (C1)--(C3) on tie-free inputs. Then there exists an $L$-Lipschitz
function $H : \mathcal{R} \to [0,1]$ such that
\begin{equation}
  Q(x) = H(\rho(x)).
  \label{eq:admissible_form}
\end{equation}
Conversely, any $Q(x) = H(\rho(x))$ with $H$ Lipschitz satisfies (C1) and (C3),
and satisfies (C2) if and only if $H$ takes no batch-dependent argument.
\end{theorem}

\begin{proof}
See Appendix~\ref{app:proofs}.
\end{proof}

\begin{proposition}[Stability bound]
\label{prop:stability}
If $Q(x) = H(\rho(x))$ with $H$ $L_H$-Lipschitz, then
$|Q(x) - Q(x')| \le L_H\,\norm{\rho(x) - \rho(x')}$ for all $x, x'$.
\end{proposition}

This is immediate from the Lipschitz property of $H$. The Lipschitz constant of $H$
controls sensitivity directly, unlike differentiable sorting operators whose
sensitivity to near-tie inputs grows with the temperature.
\section{Discussion}
\label{sec:conclusion}
SoftSort-type relaxations violate (C1) by a positive explicit margin, and batchwise
rank operators violate (C2) by a margin approaching 1. Condition (C3) implies (C1)
and should not be read as a third independent failure mode.
Theorem~\ref{thm:structural_form} identifies the admissible class with operators
that factor through $\rho$ via a Lipschitz function and carry no batch dependence.
This distinction remains relevant even as recent work proposes more unified
differentiable order-type operators for ranking, sorting, and top-$k$
selection~\citep{Struski2025}. Such advances concern optimization and computational
surrogacy, while admissibility here is defined by ordinal invariance and batch
independence.

Conditions (C1)--(C3) define one design space for rank normalization.
Other settings, such as those permitting limited batch dependence or featurewise
across-sample normalization, require separate axiomatic treatment. The violation
bounds are constructive but not tight. Whether admissibility has implications for
generalization bounds in rank-based learning is an open question.

\appendix
\section{Proofs}
\label{app:proofs}
\begin{lemma}
\label{lem:c1_factorization}
If $Q$ satisfies (C1), then $Q(x \mid B) = Q(x' \mid B)$ whenever
$x \sim_{\mathrm{ord}} x'$. Consequently, if $Q$ also satisfies (C2), there exists
a well-defined $H : \mathcal{R} \to [0,1]$ such that $Q(x) = H(\rho(x))$ for all
tie-free $x$.
\end{lemma}

\begin{proof}
Let $x \sim_{\mathrm{ord}} x'$. Enumerate the distinct values of $x$ as
$v_1 < \cdots < v_m$ and those of $x'$ as $v'_1 < \cdots < v'_m$; the cardinalities
agree because $x \sim_{\mathrm{ord}} x'$ implies the same pairwise ordering. Define
$g(v_k) = v'_k$ for $k = 1,\ldots,m$, extended linearly on each interval
$(v_k, v_{k+1})$ and with slope $1$ outside $[v_1, v_m]$. Then $g$ is strictly
increasing, $g^{\otimes d}(x) = x'$, and (C1) gives
$Q(x' \mid B) = Q(g^{\otimes d}(x) \mid B) = Q(x \mid B)$.

Since $Q$ is constant on order-equivalence classes and $\rho$ is a complete
invariant of those classes, the assignment $H(\rho(x)) := Q(x)$ is well-defined
once (C2) removes the batch argument.
\end{proof}

\begin{lemma}
\label{lem:c23_H}
If $Q(x) = H(\rho(x))$ satisfies (C2), then $H$ depends on no batch-derived
quantity. If $Q$ further satisfies (C3), then $H$ is $L$-Lipschitz on $\mathcal{R}$.
\end{lemma}

\begin{proof}
For (C2): if $H$ depended on a batch-derived quantity $u(B)$, then
$Q(x \mid B) = \tilde{H}(\rho(x), u(B))$. Condition (C2) requires
$\tilde{H}(\rho(x), u(B_1)) = \tilde{H}(\rho(x), u(B_2))$ for all $B_1, B_2$
containing $x$. Since $u(B)$ ranges over arbitrary values, $\tilde{H}$ must be
independent of its second argument.

For (C3): the bound $|Q(x) - Q(x')| \le L\norm{\rho(x) - \rho(x')}$ translates
directly to $|H(r) - H(r')| \le L\norm{r - r'}$ for all $r, r' \in \mathcal{R}$,
so $H$ is $L$-Lipschitz.
\end{proof}

\begin{proof}[Proof of Theorem~\ref{thm:structural_form}]
The forward direction follows from Lemmas~\ref{lem:c1_factorization}
and~\ref{lem:c23_H}: (C1) gives $Q(x) = H(\rho(x))$, (C2) removes batch dependence,
and (C3) gives the Lipschitz bound on $H$.

For the converse, let $Q(x) = H(\rho(x))$ with $H$ Lipschitz and no batch argument.
(C1) holds because $\rho(g^{\otimes d}(x)) = \rho(x)$ by definition of $\rho$.
(C3) holds because $|Q(x) - Q(x')| \le L_H\norm{\rho(x) - \rho(x')}$.
(C2) holds because $Q$ takes no batch-dependent argument.
\end{proof}

\end{document}